# Bidirectional Self-Training with Multiple Anisotropic Prototypes for Domain Adaptive Semantic Segmentation

Yulei Lu
Zhejiang University
Hangzhou, China
luyvlei@163.com

Yawei Luo*
Zhejiang University
Hangzhou, China
yaweiluo329@gmail.com

Li Zhang
Zhejiang Insigma Digital Technology
Co., Ltd.
Hangzhou, China
zhangli@insigma.com.cn

Zheyang Li
Hikvision Research Institute
Hangzhou, China
lizheyang@hikvision.com

Yi Yang
Zhejiang University
Hangzhou, China
yangyics@zju.edu.cn

Jun Xiao
Zhejiang University
Hangzhou, China
junx@cs.zju.edu.cn

## ABSTRACT

A thriving trend for domain adaptive segmentation endeavors to generate the high-quality pseudo labels for target domain and retrain the segmentor on them. Under this self-training paradigm, some competitive methods have sought to the latent-space information, which establishes the feature centroids (*a.k.a* prototypes) of the semantic classes and determines the pseudo label candidates by their distances from these centroids. In this paper, we argue that the latent space contains more information to be exploited thus taking one step further to capitalize on it. Firstly, instead of merely using the source-domain prototypes to determine the target pseudo labels as most of the traditional methods do, we bidirectionally produce the target-domain prototypes to degrade those source features which might be too hard or disturbed for the adaptation. Secondly, existing attempts simply model each category as a *single* and *isotropic* prototype while ignoring the variance of the feature distribution, which could lead to the confusion of similar categories. To cope with this issue, we propose to represent each category with *multiple* and *anisotropic* prototypes via Gaussian Mixture Model, in order to fit the *de facto* distribution of source domain and estimate the likelihood of target samples based on the probability density. We apply our method on GTA5->Cityscapes and Synthia->Cityscapes tasks and achieve 61.2% and 62.8% respectively in terms of mean IoU, substantially outperforming other competitive self-training methods. Noticeably, in some categories which severely suffer from the categorical confusion such as "truck" and "bus", our method achieves 56.4% and 68.8% respectively, which further demonstrates the effectiveness of our design. The code and model are available at https://github.com/luyvlei/BiSMAPs.

*Yawei Luo is the corresponding author

## 1 INTRODUCTION

Semantic segmentation is a fine-grained image understanding task with the goal of assigning a specific category to each pixel. Recently, this task has achieved remarkable progress with the development of deep neural network [1, 3–5, 18, 42]. The satisfying performance, nevertheless, usually comes with a price of expensive and laborious label annotations. One of the thriving solutions to mitigate this issue has sought to the synthetic datasets rendered from simulators and game engines [28, 29]. However, the notorious domain shift [31] impedes the model trained on synthetic images to be further deployed in a practical environment. To deal with this issue, domain adaptation (DA) approaches [11, 13, 15, 16, 19–23, 26, 30, 33, 34, 43] are proposed to bridge the gap between the source and target domains. In practice, the unsupervised domain adaptation (UDA), which does not need any labeled examples from the target domain, received more attention since it minimizes human labor ultimately.

Under the UDA setting, current state-of-the-art methods [25, 26, 41, 45] usually endeavor to generate high-quality pseudo labels and transfer the UDA problem into a self-learning task. The common and pivotal steps in this campaign consist of 1) training an initial adaptation model across domains, which is also called "warmup stage" and 2) generating pseudo labels to self-train the initial model towards target domain. In the warmup stage, adversarial training [23, 26, 33, 34] and style transfer [14, 17, 24, 38] techniques are most widely used. Specifically, adversarial training utilizes a discriminator to align the distributions of different domains while style transfer converts the source-domain images into the target style to train the model. For the self-training stage, assigning pseudo labels based on the prediction confidence is a common practice, yet determining the confidence thresholds for variant classes is non-trivial. More recently, applying "feature centroids" (*a.k.a* prototypes) of the semantic classes to assign pseudo labels [41] in the latent space mitigates the issue above. Such strategy determines the pseudo label candidates by their distances from feature centroids, which outperforms the approaches of using prediction confidence and becomes a vital enabling factor of many competitive methods.

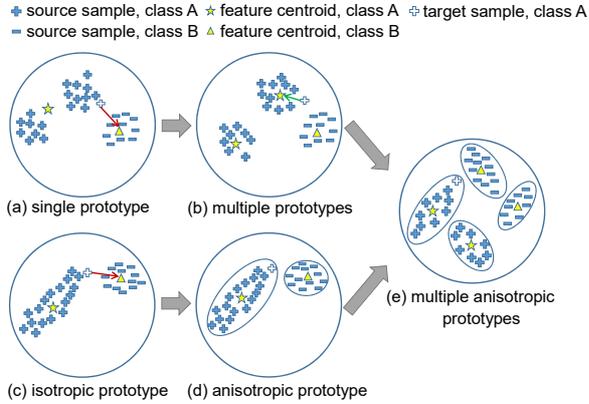

Figure 1: (Best viewed in color.) Comparison of single isotropic prototype and multiple anisotropic prototypes. (a) Traditional single prototype. Since the features of a certain category do not always obey a single-cluster cohesion in the semantic segmentation, the centroid of class A may fall into a low-density region and cause a target sample "+" to locate closer to centroid B. (b) Representing class A with multiple prototypes mitigates this issue, where the target sample "+" can be correctly classified. (c) Traditional isotropic prototype merely considers the centroid of the feature distribution. Because of the ignorance of variance, the target sample "+" is misclassified to centroid B. (d) Anisotropic prototype, where the ellipses represent contour lines of the distribution probability. Taking variance of the feature distribution into account can assign the target sample "+" correctly. (e) BiSMAP takes advantage of both (b) and (d) to model the categorical features with multiple anisotropic prototypes.

This paper follows a self-training paradigm relying on latent-space information. Upon reviewing the recent attempts along this vein, we notice several potential associated issues on the pseudo label assignment mechanism that decline the adaptation performance. **First**, existing approaches conduct the adaptation using whole source-domain information but ignore the fact that some hard and disturbed source samples do not contribute or even impede the target domain performance. For instance, the GTA5 dataset contains a vast of pixels for mountain areas while Cityscapes does not. To force the domain alignment introducing these pixels would on the contrary drift the target distribution. **Second**, traditional methods assume that each category obeys an isotropic distribution with the same variance, thus simply using a single feature centroid as the prototype and employing Euclidean distance as the metric to evaluate the similarity of a candidate feature to the current prototype. In this way, a feature that is close enough to a prototype will be assigned a pseudo label. Nevertheless, this assumption does not necessarily hold for the pixel-level features of semantic segmentation. For example, the category "vegetation" is a single class but it actually includes variant parts such as trunk and crown. When using a single prototype to represent "vegetation", the features of trunk might be improperly mapped closer to the prototype of "pole", as shown in Fig. 1(a). Besides, simply using the Euclidean distance as the metric while ignoring the distinct variance of each class may further deteriorate the pseudo label assignment between similar categories, as shown in Fig. 1(c).

These observations motivate our design of a novel bidirectional self-training method with multiple anisotropic prototypes (dubbed BiSMAP). BiSMAP argues that the latent space actually contains more information to be exploited thus taking one step further to capitalize on it. Firstly, instead of merely using the source-domain prototypes to determine the target pseudo labels as most of the traditional methods do, BiSMAP bidirectionally produces the target-domain prototypes to degrade those source features which might be too hard or disturbed for self-training. Secondly, instead of simply modeling each category as a *single* and *isotropic* prototype while ignoring the variance of the feature distribution that leads to the confusion of similar categories, BiSMAP proposes to represent each category with *multiple* and *anisotropic* prototypes via Gaussian Mixture Model, in order to fit the *de facto* distribution of source domain and estimate the likelihood of target samples based on the probability density, as shown in Fig. 1(e). In this manner, each feature can be assigned to the correct class more accurately during the adaptation.

We apply BiSMAP on GTA5->Cityscapes and Synthia->Cityscapes tasks and achieve 61.2% and 62.8% respectively in terms of mean IoU, substantially outperforming other competitive self-training methods. Noticeably, in some categories which severely suffer from the categorical confusion such as "truck" and "bus", BiSMAP achieves 56.4% and 68.8% respectively, which further demonstrates the effectiveness of our design.

## 2 RELATED WORKS

This section will review the existing works on Unsupervised Domain Adaptation, Self-training, and Gaussian Mixture Model techniques, respectively.

**Unsupervised Domain Adaptation**. To deal with the performance gap between domains, numerous works have been explored to align source and target distributions. Feature alignment based on adversarial training is preferred for UDA of segmentation tasks, which use a discriminator to guide the model to generate domain-invariant features [23, 33, 34]. In addition to the global feature alignment, category level alignment based on prototype [41] has emerged as another solution, which directly decreases the Euclidean distance between source and target features. Motivated by the recent image-to-image translation works, some works employ the style transfer technique to alleviate style differences of images from different domains, thus reducing the domain gap before training [13, 17, 38].

**Self-training**. Self-training is a semi-supervised learning method which offers competitive performance for UDA. These methods first train an initial adaptation model across domains and then relies on the model predictions to assign pseudo labels. Assigning pseudo-labels of high prediction confidence is a common practice of self-training. The pixels of common categories tend to be high confidence which leads to the rare categories can't be assigned pseudo labels, resulting in the model bias towards easy categories and thus ruining the performance of the rare ones. Nevertheless, it suffers from the noise of pseudo labels since pseudo labels with high confidence might not always be correct. Recent pseudo-label selection methods are developed to deal with the above problem. Zou *et al.* [25, 45] try to find appropriate thresholds to generate pseudo labels of rare categories under the principle of class balance. Pan *et al.* [26] use the image-level entropy to split the target domain data into two groups, then select the group of lower entropy for self-training. Pseudo-label assignment based on the class prototype is another way. Zhang *et al.* [41] use the category centroid as a prototype to assign pseudo labels, which could mitigate the side effect of unbalanced data distribution and dig up more samples for self-training. Zhang *et al.* [40] dynamically update the pseudo labels to correct the wrong labels, producing better results.

**Gaussian Mixture Model**. Gaussian Mixture Model (GMM) [27] is a probabilistic model that represents the presence of subpopulations within an overall population. A typical scenario of GMM is clustering and density estimation. Wang *et al.* [36] cluster the entropy of each class to divide the unlabeled samples into two groups, so as to find the reliable pseudo labels. Zong *et al.* [44] handles the density estimation problem in anomaly detection based on GMM. It can also work as a prototype. Yang *et al.* [37] utilizes GMM as multiple prototypes to alleviate the semantic ambiguity caused by single prototype in few-shot learning.

## 3 METHOD

### 3.1 Problem Setting

First, we formally introduce the problem of domain adaptive semantic segmentation. Under this setting, we are given a source dataset with full annotation $\{X^s, Y^s\}$ and an unlabeled target dataset $\{X^t\}$, both of which share the same category set $C$. The goal is to utilize these two datasets to train a segmentor that can apply to the target domain. In general, the segmentor $G$ can be divided into a feature extractor $E$ and a classifier $F$:

$$G = F \circ E. \tag{1}$$

The source-domain knowledge can be learned under a typical supervised learning paradigm:

$$L_{seg} = \mathbb{E}_{x \sim X^s, y \sim Y^s}[\ell(G(x), y], \tag{2}$$

where $\ell(.,.)$ denotes a proper loss function such as cross entropy. In the target domain, self-training methods tend to generate the pseudo labels of the target samples $\{\hat{Y}^t\}$ first and retrain the model on $\{X^t, \hat{Y}^t\}$ together with $\{X^s, Y^s\}$ reusing Eq. 2.

However, as mentioned above, existing attempts in latent space usually ignore the variance and multi-cluster trait of the feature distribution, thus generating the noisy $\hat{Y}^t$. Additionally, some hard or disturbed samples in $X^s$ are enrolled in the training process, which would impede the adaptation performance. To deal with these issues, we propose the BiSMAP method. On one hand, it utilizes the "Multiple Anisotropic Prototypes" (MAPs) to generate more accurate pseudo labels for the retraining phase. On the other hand, it bidirectionally introduces the target-domain prototypes to degrade those hard or disturbed source samples. These designs will be detailed in the following.

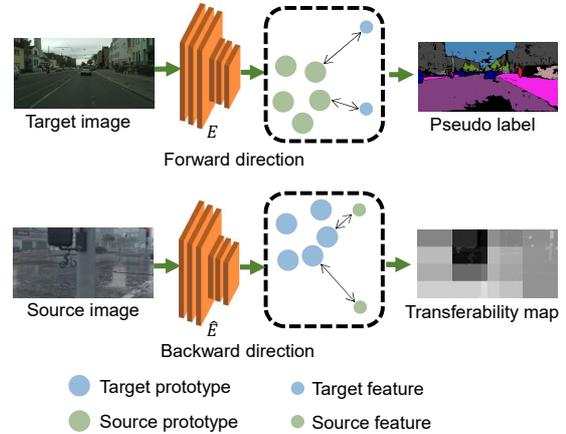

Figure 2: (Best viewed in color.) The schematic diagram of BiSMAP, in which the prototypes from both domains are innovatively introduced. In the "forward direction", we utilize the proposed MAPs to do the pseudo labels selection. These high-quality pseudo labels are then leveraged to retrain the segmentor. In the proposed "backward direction", we retrain the segmentor with both source data and selected target pseudo labels. In this process, we employ the target-domain prototypes to generate STM and reweight the pixel-level training loss on source samples with STM (*i.e.*, the source pixels in the small-value areas of the STM will be degraded in term of their training losses).

### 3.2 Bidirectional Self-training

A brief overview of our bidirectional idea is illustrated in Fig. 2. It consists of a "forward direction" that selects reliable pseudo labels in the target domain resorting to the source-domain prototypes, as well as a complementary "backward direction" to degrade those hard or disturbed source samples according to their relations to the target-domain prototypes. In the forward direction, we improve the traditional self-training methods by introducing the "Multiple Anisotropic Prototypes" to generate more accurate pseudo labels. In the backward direction, we generate the Source Transferability Map (STM) to represent the importance of each source sample in the adaptation and accordingly reweight the training loss map. For convenience, we begin with the introduction of MAPs.

### 3.3 Multiple Anisotropic Prototypes

As analyzed above, the *de facto* feature distribution of a category is exhibited as multi-cluster and anisotropic, which goes beyond the representation ability of a traditional single centroid. To address

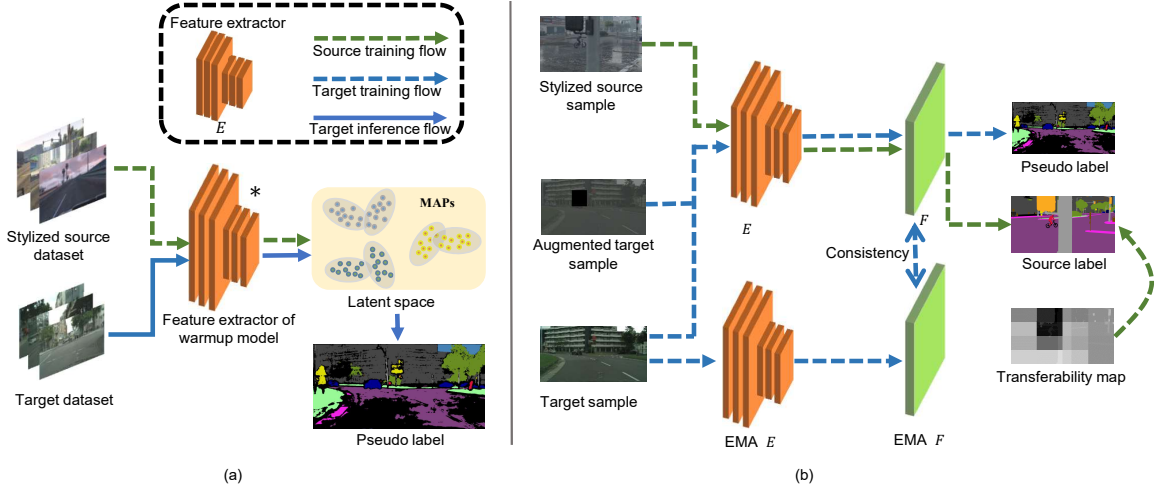

Figure 3: The architecture of BiSMAP. (a) Pseudo label generation phase. The feature extractor $E$ is pre-trained using the stylized source dataset in the previous warmup stage and kept fixed in this phase. In the source training flow, source features extracted by $E$ are clustered by GMM in a class-ware mode to generate MAPs. In the target inference flow, the target features extracted by $E$ are evaluated by MAPs and assigned high-quality pseudo labels. (b) Retraining phase. $E$ is retrained with the source samples together with the target pseudo labels generated in the first phase. During retraining, the loss of the source samples is reweighted by the source transferability map (STM). To further mitigate the noise of pseudo labels, consistency regularization based on the exponential moving average (EMA) model and data augmentation is applied in this process.

this limitation, we propose Multiple Anisotropic Prototypes (MAPs), as illustrated in Fig. 3(a).

**MAPs Generation using Gaussian Mixture Model**. To produce MAPs of the source domain, we match the source feature distribution with a Gaussian Mixture Model (GMM). GMM is a classical model which can fit the complex distributions and estimate the probability density [27]. The advantage of choosing GMM is multi-fold. On one hand, GMM employs probability density instead of the isotropic distance to measure the probability that a sample belongs to a cluster. On the other hand, GMM harnesses multiple weighted Gaussian distributions and hence can be extended to non-Gaussian cases. To get MAPs of each category, we leverage the encoder $E$ to extract the features of each class on the source domain. Specifically, only those source features that are correctly classified are considered in this process:

$$\Lambda_c = \{f_i^s \mid \arg\max G(x^s)_i = c, y_i^s = c, x^s \in X^s\}, \quad (3)$$

where $f_i^s$ is calculated by $E(x^s)_i$ and represents the feature extracted by $E$ at pixel index[1] $i$ of $x^s$, $c$ denotes a specific category, $x^s$ and $y^s$ are the source image and its corresponding label. $E$ denotes the feature extractor of the segmentor which is trained by domain alignment methods, e.g., adversarial training, or style transfer. $\Lambda_c$ denotes the feature set of category $c$ in the source domain.

GMM works in a class-aware mode in which the features $\Lambda_c$ from each category are fitted by a specific GMM. In this manner, we can get $C$ GMMs corresponding to $C$ semantic category. Each

GMM takes the form of a weighted sum of a series of Gaussian distributions given by:

$$p_c(f) = \log(\sum_{k=1}^{K} \pi_{c,k} N_{c,k}(f \mid \mu_{c,k}, \Sigma_{c,k})), \quad (4)$$

$$N_{c,k}(f) = \frac{\exp\{-\frac{1}{2}(f - \mu_{c,k})^T \Sigma_{c,k}^{-1}(f - \mu_{c,k})\}}{(2\pi)^{d/2} |\Sigma_{c,k}|^{1/2}}, \quad (5)$$

where $f$ is a feature vector, $\mu$ and $\Sigma$ are the mean vector and covariance matrix of the Gaussian distribution, $K$ represents the number of Gaussian distributions, $d$ is the dimension of feature vector, and $\pi$ denotes the mixture weight. $c$ and $k$ are the category index and the prototype index, respectively. Each category has $K$ prototypes and each prototype follows an anisotropic Gaussian distribution. For the feature set $\Lambda_c$ of a specific category $c$ and the initial prototype $p_c$, we can apply Expectation-Maximization algorithm [27] to solve the GMM equation iteratively and get the prototype $p_c$ composed of GMM. Since each category has $K$ prototypes and each prototype has its covariance matrix, our design breaks the constraints of the improper "single-centroid and isotropic" distribution assumption.

Note that due to the heavy memory footprint of the EM algorithm facing the large-scale pixel-level samples, it is impractical to apply EM straightly on the whole source-domain features. Instead, we sample 300, 000 features from each category, compromising to the scale that an EM algorithm can handle efficiently.

**Feature Similarity Measurement**. We take the prototype $p_c$ corresponding to category $c$ as an example to illustrate the measurement of similarity between target-domain features and source-domain prototypes. Given a target image $x^t$, we use the feature extractor $E$ to get the feature map $f^t = E(x^t)$. For any given pixel-level feature $f_i^t$, we can measure its similarity to the prototype as $p_c(f_i^t)$. Such similarity is based on the probability density rather than the isotropic distance so that it can cope with more complex distributions, and we call $\{p_c \mid c \in C\}$ multiple anisotropic prototypes. Therefore, for each pixel feature we can get a set of probabilities $\{p_c(f_i^t) \mid c \in C\}$.

---
[1] Unless otherwise noted, we upsample the feature map to the same size of the input image when generating the transferability map and pseudo labels. In this way, a pixel index $i$ can refer to the same position in both feature map and input image.

**Pseudo Label Assignment**. MAPs can estimate the probability of generating a sample of a certain category at any location in latent space. With the probability between each pixel-level feature and the prototypes, we assign pseudo labels based on the log probability density. We take a specific pixel feature $f_i^t$ for example. First, we get the similarity set $\{p_c(f_i^t) \mid c \in C\}$ using Eq. 4. Then we define the pseudo label of pixel $i$ in target image $x^t$ as $\hat{y}_i^t$, which is is a one-hot vector or an all-zero vector. The $r$-th bit of $\hat{y}_i$ is calculated by the following formula:

$$\hat{y}_{ir}^t = \begin{cases} 1, & r = \arg\max_c (p_c(f_i^t)), p_r(f_i^t) \geq \delta \\ 0, & otherwise. \end{cases} \quad (6)$$

Our pseudo label assignment strategy is based on probability density: samples with log probability density higher than $\delta$ will be assigned pseudo labels and others will be ignored. The target domain dataset with pseudo labels denotes as $\{X^t, \hat{Y}^t\}$.

### 3.4 Source Sample Degradation

In traditional self-training methods, some hard or disturbed samples in $X^s$ are easily enrolled in the training process, which would impede the adaptation performance. In this section, we would filter out these samples by estimating their transferability and reducing their influence on adaptation.

**Target Prototypes Generation via Clustering.** As shown in Fig. 2, we measure the transferability of each source sample with the aid of target-domain prototypes. Due to the unlabeled characteristic of the target domain, we are not able to use any categorical information to generate the prototypes like traditional methods. Accordingly, a clustering-based method is proposed. First, we employed a network $\hat{E}$ pre-trained on a third-party dataset, e.g., ImageNet, to extract semantic features from the target domain and get a feature set. Then an unsupervised clustering algorithm, e.g., K-Means [10], is applied to cluster these features. Finally, we keep the cluster centers $\{A_j \mid 0 < j < J\}$ as the target prototypes where $J$ is the number of clusters used in K-Means. These prototypes are later employed to evaluate the transferability of source domain features, as shown in Fig. 2.

**Source Transferability Map.** Now we have gotten a set of prototypes that could represent the feature distribution of the target domain, in the next step we will evaluate the transferability of source features and figure out those hard or disturbed samples. Here we propose to estimate the transferability of each sample based on both feature distance and category entropy. Formally, we first define the distance from the source features to the target domain as follows:

$$D(x^s)_i = min\{\left\|\hat{f}_i^s - A_j\right\|_2 \mid 0 < j < J\}, \quad (7)$$

where $\hat{f}^s$ represents the feature of $x^s$ extracted by $\hat{E}$, $A_j$ is one of the target prototypes and $i$ is the pixel index on feature map $\hat{f}^s$. $D(x^s)_i$ coarsely depicts the transferability of a pixel-level source feature but suffers from the long-tailed distribution: we find the samples from head classes can always got a relatively high score in the transferability map and vise versa. To balance the class-level transferability score, we calculate the average entropy of the predicted results for each class in the target domain using the source pre-trained segmentation model. Then we apply Min-Max normalization to those category entropy:

$$e_c' = \frac{e_c - e_{min}}{e_{max} - e_{min}}, \quad (8)$$

where $e_c$ denotes the mean entropy of category $c$ in the target domain, $e_{min}$ and $e_{max}$ represent the maximum and minimum entropy of all categories, respectively. $e_c$ is calculated from the softmax layer of the source pre-trained segmentation model on target domain. Since there are no labels for the target domain, we use pseudo-labels to calculate the category entropy. Finally, we define the transferability map of $x^s$ as the follows:

$$w_i^s = min\{\exp(\frac{-D(x^s)_i^2}{d_{mean}^2}\log 2) + e_c', 1.0\}, \quad (9)$$

where $d_{mean}$ denotes the mean distance calculated by $D(\cdot)$ in the whole source dataset, $w^s$ is the transferability map and $w_i^s$ represents the value of $w^s$ at pixel index $i$ and $c$ is the category of $x_i^s$ according to $y_i^s$. Obviously, Eq. 9 contains a term based on feature distance and a term based on category entropy. The feature distance determines the overall transferability of a source region, while the category entropy serves as a lower bound to balance the class-level transferability. The shape of the transferability map $w^s$ is consistent with the shape of the source images, as shown by Fig. 2. We will detail the usage of $w^s$ during training in Section 3.5.

### 3.5 Training Pipeline

The proposed BiSMAP is composed of a warmup stage and a self-training stage, and the latter can be further divided into a pseudo label generation phase and a retraining phase. We warm up the initial model with Eq. 2, in which we stylize the source images using Global Photometric Alignment [24] technique following the setting of [24][2]. These stylized source data will be also employed in the self-training stage.

In MAPs generation described in Sec. 3.3, the model pre-trained in the warmup stage is applied as the feature extractor $E$ for producing MAPs. We sample 300, 000 features per class and implement GMM with 8 diagonal covariance Gaussian distributions. Then the pseudo labels are generated via MAPs according to Eq. 6. For STM generation described in Sec. 3.4, we employ a ResNet152 pre-trained by SimCLRv2 [6] as the feature extractor $\hat{E}$.

The proposed BiSMAP is featured by three loss functions, i.e., the reweighted source loss, the target self-training loss, and the consistency regularization loss. Given an image $x^s \in X^s$ of shape $3 \times H \times W$ and a label map $y^s \in Y^s$ of shape $C \times H \times W$ where $C$ is the number of semantic classes, the loss of each sample in source

---
[2]GPA is also served as our baseline method.

Table 1: Comparison of domain adaption tasks in GTA to Cityscapes. "*" indicates the results after the distillation stage.

| Methods | road | side. | buil. | wall | fence | pole | light | sign | vege. | terr. | sky | pers. | rider | car | truck | bus | train | motor | bike | mIoU |
|---|---|---|---|---|---|---|---|---|---|---|---|---|---|---|---|---|---|---|---|---|
| CAG_UDA [41] | 90.4 | 51.6 | 83.8 | 34.2 | 27.8 | 38.4 | 25.3 | 48.4 | 85.4 | 38.2 | 78.1 | 58.6 | 34.6 | 84.7 | 21.9 | 42.7 | 41.1 | 29.3 | 37.2 | 50.2 |
| IAST [25] | 94.1 | 58.8 | 85.4 | 39.7 | 29.2 | 25.1 | 43.1 | 34.2 | 84.8 | 34.6 | 88.7 | 62.7 | 30.3 | 87.6 | 42.3 | 50.3 | 24.7 | 35.2 | 40.2 | 52.2 |
| FDA [38] | 92.5 | 53.3 | 82.4 | 26.5 | 27.6 | 36.4 | 40.6 | 38.9 | 82.3 | 39.8 | 78.0 | 62.6 | 34.4 | 84.9 | 34.1 | 53.1 | 16.9 | 27.7 | 46.4 | 50.5 |
| Coarse-to-fine [24] | 92.5 | 58.3 | 86.5 | 27.4 | 28.8 | 38.1 | 46.7 | 42.5 | 85.4 | 38.4 | 91.8 | 66.4 | 37.0 | 87.8 | 40.7 | 52.4 | 44.6 | 41.7 | 59.0 | 56.1 |
| ProDA [40] | 91.5 | 52.4 | 82.9 | 42.0 | 35.7 | 40.0 | 44.4 | 43.3 | 87.0 | 43.8 | 79.5 | 66.5 | 31.4 | 86.7 | 41.1 | 52.5 | 0.0 | 45.4 | 53.8 | 53.7 |
| MFA [39] | **94.5** | **61.1** | **87.6** | 41.4 | 35.4 | 41.2 | 47.1 | 45.7 | 86.6 | 36.6 | 87.0 | **70.1** | 38.3 | 87.2 | 39.5 | 54.7 | 0.3 | 45.4 | 57.7 | 55.7 |
| GPA(baseline) | 76.8 | 34.6 | 68.2 | 22.7 | 21.4 | 40.1 | 44.1 | 26.5 | 85.4 | 29.4 | 74.6 | 67.4 | 27.6 | 87.9 | 37.7 | 47.5 | 34.3 | 29.2 | 24.7 | 46.3 |
| BiSMAP(ours) | 86.2 | 48.4 | 83.5 | **43.8** | **38.2** | **41.8** | **49.5** | **54.7** | **87.9** | 41.7 | 84.7 | 63.9 | 34.4 | **89.1** | **49.1** | **62.2** | 43.8 | 37.1 | 56.6 | **57.7** |
| ProDA* [40] | 87.8 | 56.0 | 79.7 | 46.3 | 44.8 | 45.6 | 53.5 | 53.5 | 88.6 | 45.2 | 82.1 | 70.7 | 39.2 | 88.8 | 45.5 | 59.4 | 1.0 | 48.9 | 56.4 | 57.5 |
| MFA* [39] | 93.5 | 61.6 | 87.0 | 49.1 | 41.3 | 46.1 | 53.5 | 53.9 | 88.2 | 42.1 | 85.8 | 71.5 | 37.9 | 88.8 | 40.1 | 54.7 | 0.0 | 48.2 | 62.8 | 58.2 |
| BiSMAP*(ours) | 89.2 | 54.9 | 84.4 | 44.1 | 39.3 | 41.6 | 53.9 | 53.5 | 88.4 | 45.1 | 82.3 | 69.4 | 41.8 | 90.4 | 56.4 | 68.8 | 51.2 | 47.8 | 60.4 | **61.2** |

Table 2: Comparison of domain adaption tasks in SYNTHIA to Cityscapes. "*" indicates the results after the distillation stage.

| Methods | road | side. | buil. | light | sign | vege. | sky | pers. | rider | car | bus | motor | bike | mIoU |
|---|---|---|---|---|---|---|---|---|---|---|---|---|---|---|
| CAG_UDA [41] | 84.7 | 40.8 | 81.7 | 13.3 | 22.7 | 84.5 | 77.6 | 64.2 | 27.8 | 80.9 | 19.7 | 22.7 | 48.3 | 51.5 |
| IAST [25] | 81.9 | 41.5 | 83.3 | 30.9 | 28.8 | 83.4 | 85.0 | 65.5 | 30.8 | 86.5 | 38.2 | 33.1 | 52.7 | 57.0 |
| FDA [38] | 79.3 | 35.0 | 73.2 | 19.9 | 24.0 | 61.7 | 82.6 | 61.4 | 31.1 | 83.9 | 40.8 | 38.4 | 51.1 | 52.5 |
| Coarse-to-fine [24] | 75.7 | 30.0 | 81.9 | 18.0 | 32.7 | 86.2 | **90.1** | 65.1 | 33.2 | 83.3 | 36.5 | 35.3 | 54.3 | 55.5 |
| ProDA [40] | **87.1** | **44.0** | 83.2 | **45.8** | 34.2 | **86.7** | 81.3 | 68.4 | 22.1 | **87.7** | **50.0** | 31.4 | 38.6 | 58.5 |
| MFA [39] | 85.4 | 41.9 | 84.1 | 22.2 | 23.9 | 83.6 | 80.7 | **71.5** | 35.8 | 86.6 | 47.6 | 37.2 | **62.5** | 58.7 |
| GPA(baseline) | 75.3 | 31.3 | 78.7 | 22.4 | 25.2 | 75.9 | 82.0 | 64.3 | 28.3 | 80.2 | 25.9 | 24.6 | 27.1 | 49.3 |
| BiSMAP(ours) | 79.1 | 36.6 | **84.7** | 31.8 | **41.2** | 84.7 | 89.2 | 65.4 | **39.0** | 79.0 | 47.6 | **41.8** | 61.9 | **60.1** |
| ProDA* [40] | 87.8 | 45.7 | 84.6 | 54.6 | 37.0 | 88.1 | 84.4 | 74.2 | 24.3 | 88.2 | 51.1 | 40.5 | 45.6 | 62.0 |
| MFA* [39] | 81.8 | 40.2 | 85.3 | 38.0 | 33.9 | 82.3 | 82.0 | 73.7 | 41.1 | 87.8 | 56.6 | 46.3 | 63.8 | 62.5 |
| BiSMAP*(ours) | 81.9 | 39.8 | 84.2 | 41.7 | 46.1 | 83.4 | 88.7 | 69.2 | 39.3 | 80.7 | 51.0 | 51.2 | 58.8 | **62.8** |

domain is reweighted according to $\{w^s\}$ during training in the form of multiplication:

$$\mathcal{L}_{ce}(x^s, \theta_G) = \sum_{i=1}^{H \times W} \sum_{c=1}^{C} -w_i^s y_{ic}^s \log p_{ic}, \quad (10)$$

where $p_{ic}$ represents the probability of class $c$ on pixel $i$. $y_{ic}^s$ is the ground truth of class $c$ on the pixel $i$. $w_i^s$ denotes the value of pixel $i$ in the transferability map illustrated in Section 3.4. $\theta_G$ is the parameter of the model $G$ to be optimized.

For the images $x^t \in \{X^t\}$ from the target domain and its corresponding pseudo label $\hat{y}^t$, we train the model with a symmetric cross-entropy loss [35]:

$$\mathcal{L}_{sce}(x^t, \theta_G) = \sum_{i=1}^{H \times W} \sum_{c=1}^{C} -\alpha \hat{y}_{ic}^t \log p_{ic} - \beta p_{ic} \log \hat{y}_{ic}^t, \quad (11)$$

We clamp the one-hot label $\hat{y}^t$ to $[1e-4, 1]$ to avoid the numerical issue of log 0. The $\alpha$ and $\beta$ are set to 0.1 and 1.0 respectively.

To further mitigate the noise in the candidate pseudo labels, we apply consistency regularization with KLD loss on the target domain:

$$\mathcal{L}_{consist}(x^t, \varphi(x^t), \theta_G) = \sum_{i=1}^{H \times W} \sum_{c=1}^{C} \widetilde{p}_{ic} \log \widetilde{p}_{ic} - \widetilde{p}_{ic} \log p_{ic}, \quad (12)$$

where $\varphi$ represents an image augmentation function and $\widetilde{p}$ is the output of the exponential moving average (EMA) model with $\varphi(x^t)$ as input as introduced by [32]. The overall training loss is:

$$\mathcal{L} = \mathcal{L}_{ce} + \mathcal{L}_{sce} + \lambda \mathcal{L}_{consist} \quad (13)$$

where $\lambda$ controls the relative weight of consistency regularization.

## 4 EXPERIMENT

### 4.1 Datasets

We evaluate BiSMAP together with several state-of-the-art algorithms on two synthetic-to-real domain adaptation tasks: GTA5 [28] to Cityscapes and SYNTHIA [29] to Cityscapes. GTA5 contains 24, 966 images with 1, 914 × 1, 052 resolution. SYNTHIA contains 9, 400 images with 1, 280 × 760 resolution. We use GTA5 and SYNTHIA as the source domain and Cityscapes as the target domain.

### 4.2 Implementation Details

We utilize the DeepLab-V3Plus[5] with ResNet101[12] as the segmentor following the setting of [24], and the backbone is pre-trained on ImageNet [8]. The source domain images are converted to the style of the target domain by GPA as introduced in [24], and all the following phase is based on the stylized dataset. Images are randomly scaled by ×0.5 ∼ ×1.5 and cropped to 896 × 512.

For the generation of MAPs, we use GMM with 8 Gaussian distributions. The probability density threshold $\delta$ is set to 100 for GTA5->Cityscapes and −50 for Synthia->Cityscapes. We use the standard color-jittering augmentation in both source and target domains as in [24]. In consistency regularization, we follow the setting of [32], where an EMA model works as a teacher to guide the main model. We use RandAugment [7] and CutOut [9] as image augmentation function $\varphi$. The $\lambda$ is set to 20, and the smoothing coefficient of EMA model is set to 0.999. The batch size is set to 8 for warmup and 4 for self-training, and the iteration of each stage is 90, 000. We use SGD [2]optimizer with momentum and weight-decay as 0.9 and 0.0005, respectively. The learning rate of SGD is set to 0.0005 and decayed by the poly policy with power 0.9.

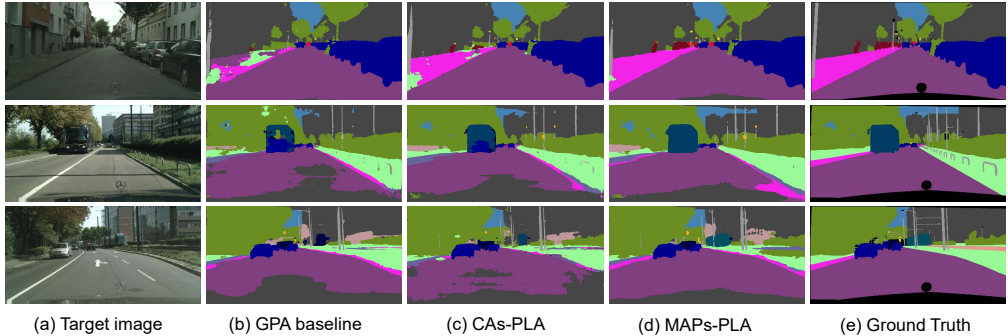

Figure 4: Qualitative results of UDA segmentation for GTA5->Cityscapes.

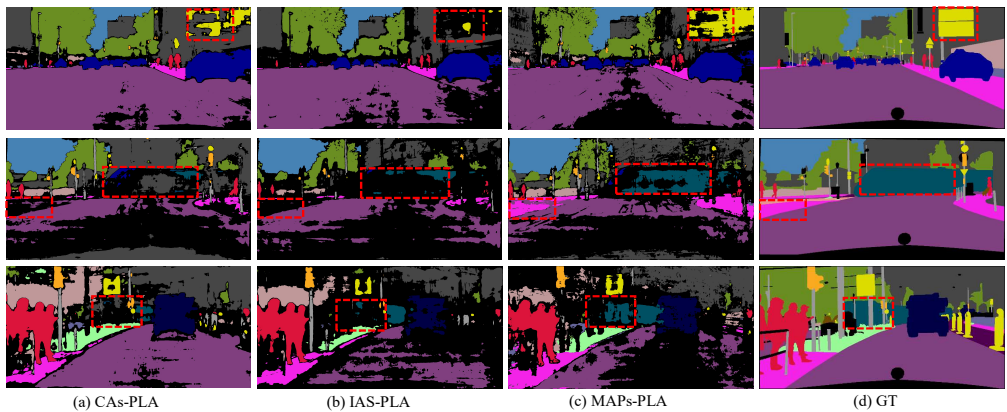

Figure 5: Pseudo label comparison to other PLA strategies.

### 4.3 Comparative Studies

**Compared with SOTA.** We evaluate BiSMAP together with several state-of-the-art methods. It's worth noting that the most of prototype-based methods are implemented on DeepLab-V3Plus since it can better corresponds a feature to a certain pixel. For fair comparison, we also report the results with an additional distillation stage which is proposed in [40]. The qualitative segmentation examples can be viewed Fig. 4.

We show the comparison results on GTA5->Cityscapes in Tab. 1. The mIoU of BisMAP after the self-training stage is 57.7%, which achieves the state-of-the-art accuracy. Besides, BiSMAP brings significant improvements to the confusion categories compared to the GPA baseline [24], e.g., truck and bus. After the distillation stage, BiSMAP achieve 61.2% mIoU, substantially outperforming the previous state of the arts [37, 39]. We show the comparison results on Synthia->Cityscapes in Tab. 2 and get a 10.8% performance gain compared to the GPA baseline. Similarly, we conduct a distillation stage and achieve SOTA 62.8% mIoU.

**Comparison of pseudo label assignment (PLA) strategies.** In Tab. 3, we compare the performance of the proposed MAPs-PLA with both prototype-based and threshold-based approaches on GTA5->Cityscapes task, including "Category Anchors pseudo label assignment" (CAs-PLA) [41] and "Instant Adaptive Selector pseudo label assignment" (IAS-PLA) [25]. All of the results are based on the same warmup model and equipped with our proposed STM for a fair comparison. CAs-PLA assigns pseudo labels using Euclidean distance to the prototypes based on the assumption that the feature distribution is isotropic and single-centered. IAS-PLA is a popular method based on the prediction confidence, but it is also significantly lower than our results.

By modeling the latent-space information more properly, MAPs-PLA boosts the mIoU with an additional 2.4% compared to CAs-PLA

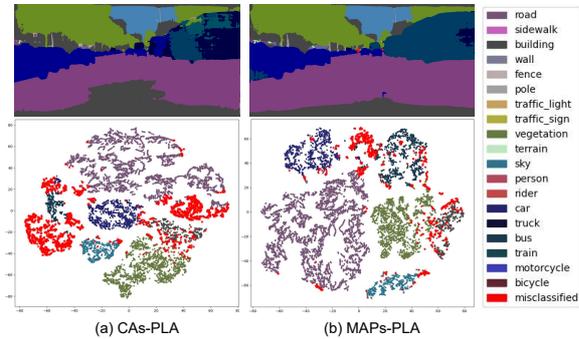

(a) CAs-PLA  (b) MAPs-PLA

Figure 6: Latent space visualization of (a) CAs-PLA and (b) MAPs-PLA using T-SNE. We use the "bright red" color to indicate those misclassified features.

Table 3: Comparison with other pseudo label assign methods.

|        | CAs-PLA | IAS-PLA | MAPs-PLA |
|--------|---------|---------|----------|
| mIoU   | 54.9    | 55.4    | **57.3** |

and 1.9% compared to IAS. We show the visualization results of different PLA strategies in Fig. 5. The region of the red dotted box in Fig. 5 shows that MAPs can not only fix some errors of CAs-PLA, but also achieve good results in pseudo-label screening. Besides, MAPs-PLA achieves less noise than CAs-PLA since the probability density function forms a more accurate estimate of the confidence.

**Comparison of T-SNE visualization results.** To further demonstrate the effectiveness of MAPs-PLA, we visualize the feature distribution trained by CAs pseudo labels and MAPs pseudo labels as shown in Fig. 6, in which the misclassified features are marked as "light red". We mainly focus on two confusing categories: trunk and bus. The single-centroid prototype confuses them since they are naturally closed in latent space, e.g., in the latent space of CAs self-trained model, the feature of bus are multiple clusters but most of the clusters are misclassified. In the latent space of MAPs self-trained model, the bus category also exhibits multi-cluster characteristics but are mostly correctly classified thanks to our Multiple Anisotropic Prototypes. We also show the visualization of latent space before self-training in Appendix. C and similar characteristics of multi-cluster can be observed. In conclusion, MAPs-PLA can better correct errors and achieve improvement in such cases.

### 4.4 Ablation Studies

To assess the importance of various aspects of BiSMAP, we investigate the effects of different components on GTA5->Cityscapes task as shown in Tab. 4. The GPA [24] baseline reaches 46.3% in our experiment. By introducing the STM, we increase the mIoU by 1.9%. And after the self-training stage with the pseudo labels generated by MAPs, the mIoU improves significantly, reaching 57.3%. BiSMAP achieves 11.4% performance improvement in one stage of self-training. After the distillation stage, we achieve 61.2% mIoU and significantly outperforms other methods.

Table 4: Ablation of each component

| baseline | STM | MAPs-PLA | consistency | distillation | mIoU |
|----------|-----|----------|-------------|--------------|------|
| ✓        |     |          |             |              | 46.3 |
| ✓        | ✓   |          |             |              | 48.2(+1.9) |
| ✓        |     | ✓        |             |              | 56.4(+10.1) |
| ✓        | ✓   | ✓        |             |              | 57.3(+11.0) |
| ✓        | ✓   | ✓        | ✓           |              | 57.7(+11.4) |
| ✓        | ✓   | ✓        | ✓           | ✓            | **61.2(+14.9)** |

Table 5: Parameter study of MAPs-PLA. For MAPs-PLA, threshold is the $\delta$. For CAs-PLA, threshold is based on spatial distance. PL ratio is the ratio of pixel assigned pseudo labels.

| method    | MAPs-PLA |      |      |      | CAs-PLA |      |      |      |
|-----------|----------|------|------|------|---------|------|------|------|
| threshold | 0        | 50   | 100  | 150  | 0.5     | 1    | 1.5  | 2    |
| PL ratio  | 87.6     | 77.5 | 61.3 | 44.3 | 82.7    | 67.8 | 54.5 | 42.2 |
| mIoU      | 56.4     | 57.1 | **57.3** | 55.7 | 53.2 | 54.9 | 53.9 | 53.9 |

### 4.5 Parameter Studies

This section investigates the sensitivity of $\delta$ and analyzes the relationship between the threshold and the ratio of pseudo labels (PL ratio). As shown in Tab. 5, the probability density threshold $\delta$ is significantly negatively correlated with the PL ratio. Therefore, more percentage of the pixel will be assigned pseudo labels with the decrease of $\delta$. When $\delta = 100$, we achieve highest mIoU by trading off the balance between the PL ratio and accuracy.

The threshold used by CAs-PLA is based on spatial distance as designed in [41]. It got highest performance 54.9% when the threshold is 1.0 and 67.8% pixels are assigned pseudo labels. However, the proposed MAPs-PLA strategy gets 55.7% when only 44.3% pixels are assigned pseudo labels. We show that, at about the same ratio of pseudo labels, MAPs-PLA achieves higher performance. Similar conclusions can be drawn at other scales of PL ratio. By comparing the segmentation results under different parameters, we can conclude that MAPs-PLA improves the robustness of pseudo labels.

## 5 CONCLUSION

In this paper, we proposed a bidirectional self-training with multiple anisotropic prototypes method for unsupervised domain adaption. Specifically, we produce the target-domain prototypes to degrade those source features and develop a robust pseudo label assignment strategy based on the multiple anisotropic prototypes. Extensive experiments verify the effectiveness and superiority of BiSMAP.

## ACKNOWLEDGMENTS

This work was supported by the National Key Research & Development Project of China (2021ZD0110700), the National Natural Science Foundation of China (U19B2043, 61976185), Zhejiang Natural Science Foundation (LR19F020002), Zhejiang Innovation Foundation(2019R52002), CCF-HIKVISION OF (20210003), and the Fundamental Research Funds for the Central Universities (226-2022-00051).

## Appendix. A  PARAMETER ANALYSIS OF K

To analyze the effectiveness of multiple prototypes, we use GMMs with different $K$ (the number of clusters) to generate pseudo-labels during training, and the results are shown in Tab. 6. We find that as $K$ increases, the performance of self-training gradually improves. When $K$ is increased to 6, the mIoU achieves 57.3%. Further increasing the number of $K$ can no longer improve performance. In conclusion, a larger $K$ value can usually stimulates a stronger GMM's ability to fit the *de facto* distribution, yet $K = 8$ is enough to match the distribution in this task. Accordingly, we choose the value of $K = 8$ in the experiment. These experimental results do not contain consistency regularization in order to pinpoint the pure effect of $K$.

Table 6: The detailed results of self-training under different parameter $K$ on GTA->Cityscapes.

| K | road | side. | buil. | wall | fence | pole | light | sign | vege. | terr. | sky | pers. | rider | car | truck | bus | train | motor | bike | mIoU |
|---|------|-------|-------|------|-------|------|-------|------|-------|-------|------|-------|-------|------|-------|------|-------|-------|------|------|
| 1 | 87.9 | 51.1 | 83.4 | 38.9 | 34.6 | 40.1 | 48.2 | 46.6 | 86.8 | 35.9 | 85.4 | 66.2 | 36.5 | 89.4 | 41.3 | 54.2 | 43.1 | 39.0 | 57.4 | 56.1 |
| 2 | 82.3 | 42.8 | 82.3 | 39.9 | 39.3 | 38.9 | 49.5 | 47.9 | 87.3 | 39.7 | 82.9 | 66.5 | 37.3 | 89.5 | 41.2 | 58.7 | 55.6 | 39.5 | 56.0 | 56.7 |
| 4 | 86.8 | 51.2 | 82.0 | 43.0 | 38.0 | 42.4 | 48.8 | 50.3 | 86.4 | 38.0 | 86.4 | 64.0 | 36.0 | 88.5 | 42.7 | 53.1 | 53.9 | 32.7 | 55.0 | 56.8 |
| 6 | 85.9 | 47.5 | 83.0 | 41.3 | 37.2 | 43.4 | 51.8 | 49.2 | 87.4 | 41.0 | 84.9 | 67.1 | 38.4 | 90.0 | 46.0 | 57.7 | 47.4 | 34.4 | 55.4 | 57.3 |
| 8 | 88.4 | 47.9 | 85.3 | 38.1 | 36.0 | 41.8 | 49.7 | 47.2 | 86.2 | 38.6 | 79.2 | 67.7 | 38.9 | 89.5 | 49.7 | 62.5 | 47.4 | 37.9 | 57.6 | 57.3 |

## Appendix. B  DETAILED ABLATION STUDY

Table 7: The detailed IoU results of ablation study on GTA->Cityscapes.

| baseline | STM | MAPs-PLA | consistency | distillation | road | side. | buil. | wall | fence | pole | light | sign | vege. | terr. | sky | pers. | rider | car | truck | bus | train | motor | bike | mIoU |
|----------|-----|----------|-------------|--------------|------|-------|-------|------|-------|------|-------|------|-------|-------|------|-------|-------|------|-------|------|-------|-------|------|------|
| ✓ | | | | | 76.8 | 34.6 | 68.2 | 22.7 | 21.4 | 40.1 | 44.1 | 26.5 | 85.4 | 29.4 | 74.6 | 67.4 | 27.6 | 87.9 | 37.7 | 47.5 | 34.3 | 29.2 | 24.7 | 46.3 |
| ✓ | ✓ | | | | 75.6 | 38.1 | 70.5 | 24.7 | 25.2 | 40.3 | 43.2 | 27.8 | 84.9 | 30.9 | 77.2 | 66.7 | 33.7 | 87.7 | 38.9 | 49.5 | 33.9 | 28.6 | 39.4 | 48.3 |
| ✓ | | ✓ | | | 87.1 | 47.1 | 84.0 | 38.0 | 39.0 | 41.5 | 49.4 | 48.2 | 86.7 | 39.0 | 79.2 | 67.7 | 37.2 | 88.7 | 43.8 | 59.9 | 42.1 | 36.2 | 57.2 | 56.4 |
| ✓ | ✓ | ✓ | | | 88.4 | 47.9 | 85.3 | 38.1 | 36.0 | 41.8 | 49.7 | 47.2 | 86.2 | 38.6 | 79.2 | 67.7 | 38.9 | 89.5 | 49.7 | 62.5 | 47.4 | 37.9 | 57.6 | 57.3 |
| ✓ | ✓ | ✓ | ✓ | | 86.2 | 48.4 | 83.5 | 43.8 | 38.2 | 41.8 | 49.5 | 54.7 | 87.9 | 41.7 | 84.7 | 63.9 | 34.4 | 89.1 | 49.1 | 62.2 | 43.8 | 37.1 | 56.6 | 57.7 |
| ✓ | ✓ | ✓ | ✓ | ✓ | 89.2 | 54.9 | 84.4 | 44.1 | 39.3 | 41.6 | 53.9 | 53.5 | 88.4 | 45.1 | 82.3 | 69.4 | 41.8 | 90.4 | 56.4 | 68.8 | 51.2 | 47.8 | 60.4 | 61.2 |

## Appendix. C  T-SNE VISUALIZATION OF LATENT SPACE BEFORE SELF-TRAINING

In Fig. 7, we show the 2D visualization results of features from "truck" and "bus". It can be seen that the features of semantic segmentation have the characteristics of large variance and multi-class clusters in the latent space.

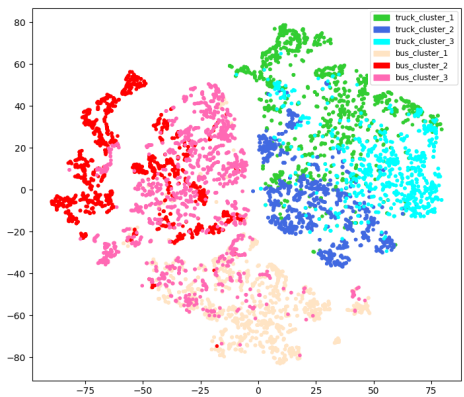

Figure 7: 2D visualization of the truck and bus features based on T-SNE. Each category shows 3 clusters obtained by GMM clustering, which are represented by different colors. All the features are extracted from GTA dataset with the warmup model.

## Appendix. D    MORE PSEUDO-LABEL COMPARISON AND QUALITATIVE RESULTS

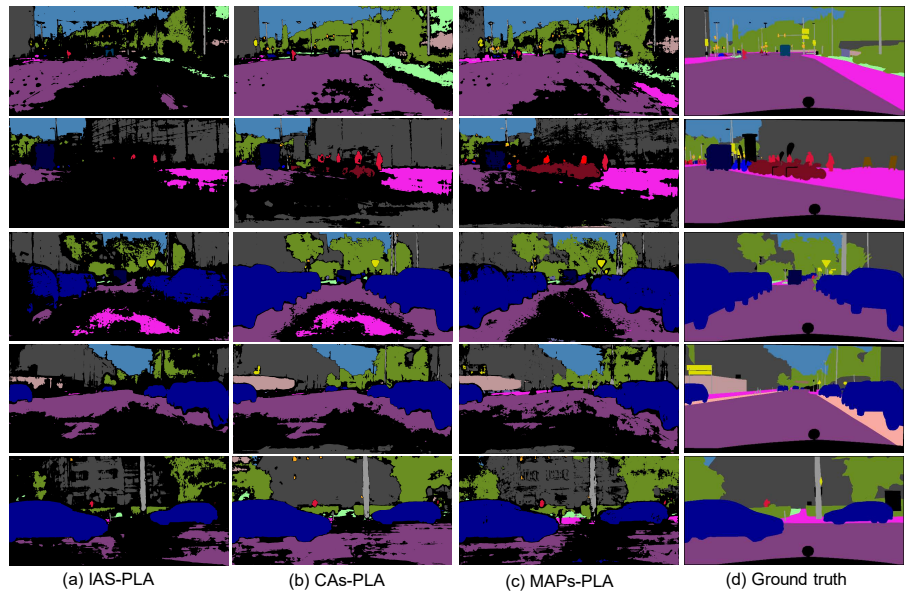

(a) IAS-PLA    (b) CAs-PLA    (c) MAPs-PLA    (d) Ground truth

**Figure 8: Pseudo label comparison with other methods: (a) IAS-PLA. (b) CAs-PLA. (c) MAPs-PLA. (d) Ground Truth.**

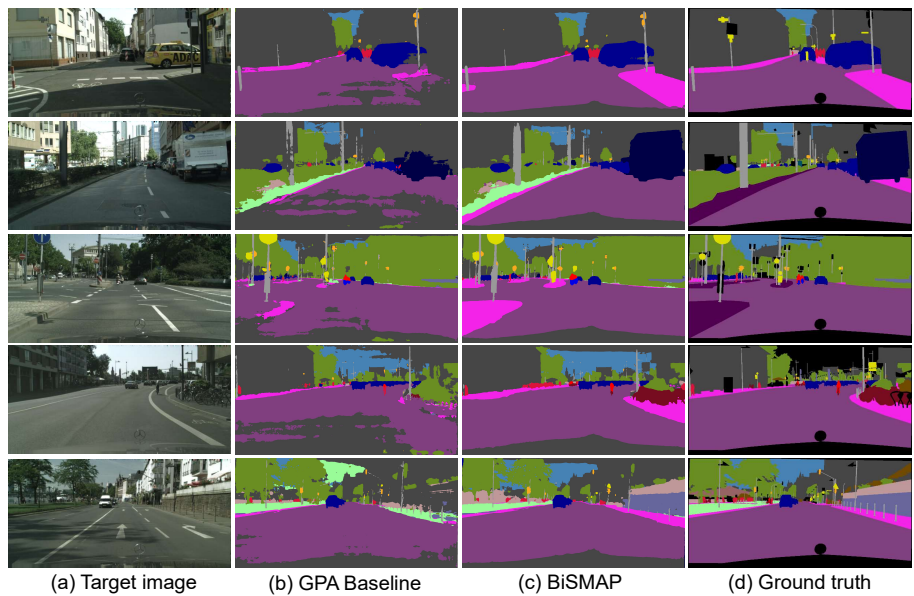

(a) Target image    (b) GPA Baseline    (c) BiSMAP    (d) Ground truth

**Figure 9: Qualitative results of UDA segmentation for GTA->Cityscapes. We visualize the results of the GPA baseline, BiSMAP (ours), and the ground truth.**